\documentclass[lettersize,journal]{IEEEtran}
\usepackage{amsmath,amsfonts}
\usepackage{algorithmic}
\usepackage{algorithm}
\usepackage{array}
\usepackage[caption=false,font=normalsize,labelfont=sf,textfont=sf]{subfig}
\usepackage{textcomp}
\usepackage{stfloats}
\usepackage{url}
\usepackage{verbatim}
\usepackage{graphicx}
\usepackage{cite}
\usepackage{booktabs}
\usepackage{multirow}
\usepackage{makecell}
\usepackage{bm}
\usepackage[]{hyperref}
\usepackage{color}
\hypersetup{hidelinks}

\begin{document}

\title{Exploiting Temporal Audio-Visual Correlation Embedding for Audio-Driven One-Shot Talking Head Animation}

\author{Zhihua Xu, Tianshui Chen, Zhijing Yang, Siyuan Peng, Keze Wang, Liang Lin, \IEEEmembership{Fellow, IEEE} 

\thanks{Zhihua Xu, Tianshui Chen, Zhijing Yang, and Siyuan Peng are with the School of Information Engineering, Guangdong University of Technology, Guangzhou 510006, China (email: \href{mailto:zihua@mail2.gdut.edu.cn}{zihua@mail2.gdut.edu.cn}, \href{mailto:tianshuichen@gmail.com}{tianshuichen@gmail.com}, \href{mailto:yzhj@gdut.edu.cn}{yzhj@gdut.edu.cn}, \href{mailto:peng0074@gdut.edu.cn}{peng0074@gdut.edu.cn}). Keze Wang and Liang Lin are with the School of Computer Science and Engineering, Sun Yat-sen University, Guangzhou 510006, China (e-mail: \href{mailto:kezewang@gmail.com}{kezewang@gmail.com}, \href{mailto:linliang@ieee.org)}{linliang@ieee.org}). (Corresponding authors: Tianshui Chen and Zhijing Yang.)}

\thanks{Manuscript received 31 January 2025; revised 2 February 2025; accepted 6 April 2025. This work was supported in part by the National Natural Science Foundation of China (NSFC) under Grant No. 62206060 and 62276283, in part by the Guangdong Basic and Applied Basic Research Foundation (Nos. 2025A1515010454, 2023A1515012561, 2023A1515012985), and in part by the Fundamental Research Funds for the Central Universities, Sun Yat-sen University under Grant 23hytd006.}

}

\markboth{IEEE TRANSACTIONS ON MULTIMEDIA,~Vol.~, No.~, August~2025}%
{Xu \MakeLowercase{\textit{et al.}}: Exploiting Temporal Audio-Visual Correlation Embedding for Audio-Driven One-Shot Talking Head Animation}


\maketitle

\begin{abstract}
The paramount challenge in audio-driven One-shot Talking Head Animation (ADOS-THA) lies in capturing subtle imperceptible changes between adjacent video frames. Inherently, the temporal relationship of adjacent audio clips is highly correlated with that of the corresponding adjacent video frames, offering supplementary information that can be pivotal for guiding and supervising talking head animations. In this work, we propose to learn audio-visual correlations and integrate the correlations to help enhance feature representation and regularize final generation by a novel Temporal Audio-Visual Correlation Embedding (TAVCE) framework. Specifically, it first learns an audio-visual temporal correlation metric, ensuring the temporal audio relationships of adjacent clips are aligned with the temporal visual relationships of corresponding adjacent video frames. Since the temporal audio relationship contains aligned information about the visual frame, we first integrate it to guide learning more representative features via a simple yet effective channel attention mechanism. During training, we also use the alignment correlations as an additional objective to supervise generating visual frames. We conduct extensive experiments on several publicly available benchmarks (i.e., HDTF, LRW, VoxCeleb1, and VoxCeleb2) to demonstrate its superiority over existing leading algorithms.
\end{abstract}


\section{Introduction}
\IEEEPARstart{A}{udio-driven} one-shot talking head animation (ADOS-THA) \cite{kim2018dvp,wang2023anyone} generates high-fidelity videos of a speaking head from a solitary reference portrait image, governed by an accompanying audio segment. It ensures the synthesized face maintains the identity captured in the static image, while concurrently facilitating the adaptation of mouth movements, head poses, and facial expressions in sync with the audio input. These techniques hold profound implications across various disciplines, including digital human animation, film production, virtual reality, and other related fields, potentially revolutionizing the way dynamic human likenesses are created and utilized in digital environments.

The principal challenges of ADOS-THA encompass the generation of lip animations that are synchronized with audio inputs while preserving natural temporal coherence. Contemporary ADOS-THA algorithms predominantly \cite{Yu2022multimodal, wang2021audio2head, yin2022styleheat, zhang2023sadtalker, Ye2023audio} focus on translating audio into semantic representations, such as facial keypoints \cite{song2021everything, Yu2022multimodal}, 3D morphable models \cite{blanz1999morphable, lahiri2021lipsync3d}, and motion models \cite{Siarohin2019fomm, wang2021audio2head}, followed by rendering techniques to produce the final visual output. However, these methods typically rely solely on the current and a few preceding audio clips, overlooking the rich audio-visual interactions and temporal correlations that could serve as valuable guidance and supervision signals. In contrast, our approach leverages a key insight: in the context of ADOS-THA, it is a mild yet reasonable assumption that the temporal relationships between adjacent audio clips strongly correlate with those of the corresponding adjacent video frames. By explicitly modeling and incorporating these cross-modal temporal correlations, our method not only achieves superior synchronization between audio and visual elements but also ensures more natural and temporally coherent results.

In this paper, we propose a temporal audio-visual correlation embedding (TAVCE) framework, which learns temporal audio-visual correlations and incorporates these correlations to bolster feature representation capabilities as well as provide additional supervision to facilitate ADOS-THA performance. Specifically, we commence by delving into the realm of audio-visual interactions, particularly focusing on the temporal correlation metric. This metric is primarily designed to identify and prioritize the inherent correlation between consecutive audio clips and their corresponding video frames. For adjacent audio clips and corresponding video frames, the metric assigns a high value. Conversely, for adjacent audio clips and non-adjacent video frames, the metric assigns a small value. Then, we incorporate the temporal correlation metric as an auxiliary objective. This not only enhances the model's proficiency in generating visual frames but also ensures that the generated frames are in harmony with their preceding audio-visual context, making the entire sequence seamless and natural. To harness its full potential, we take a step further by integrating it with the features of the current video frame. For this integration, we employ a channel attention mechanism to obtain more refined, robust, and representative features.

The contributions of this work are summarized as follows. First, we propose a temporal audio-visual correlation embedding framework, which explicitly models the temporal dependencies between audio and visual modalities to guide and regularize the generation of realistic talk head images. To our knowledge, this is the first work to introduce temporal cross-modal correlation to address this task. Second, to effectively measure and optimize the temporal relationships, we design a novel temporal audio-visual correlation metric that quantifies the synchronization and coherence between audio and visual sequences over time. This metric provides a principled way to evaluate and enhance the temporal alignment during training. Third, we introduce a channel attention mechanism to adaptively embed the temporal relationships of adjacent audio clips into visual features, enhancing the expressiveness and alignment of the generated representations. Finally, we conduct extensive experiments and present both qualitative and quantitative evaluations, demonstrating the superiority of the proposed TAVCE framework over state-of-the-art methods. The codes and trained models are available at \url{https://github.com/ZH-Xu410/TAVCE}.

\section{Related Works}
The field of one-shot talking head generation has made significant advancements, dividing into two main streams: video-driven and audio-driven techniques. Video-driven methods leverage dynamic visual cues, using sequences from existing videos to animate and create facial movements \cite{Siarohin2019fomm, wang2019few, siarohin2021motion, zhao2022thin, hong2022depth, Bounareli2023Hyper}. These approaches typically involve sophisticated deep learning models trained to replicate complex facial expressions and synchronize with spoken words \cite{chen2024learning, xu2024self, chen2025contrastive}. The visual inputs differ in format; some use 3D facial model priors, such as 3DMM \cite{blanz1999morphable}, to edit portrait images via parameter modulation \cite{geng2018warp, doukas2021headgan, ren2021pirenderer}. For example, DVP \cite{kim2018dvp} alters parameters from both the source and target videos, employing a network to render shading. Head2Head++ \cite{doukas2021head2head++} utilizes a sequential generator and a dynamics discriminator for making temporally consistent videos. An alternative approach involves mimicking movements from 2D images instead of manipulating 3D model parameters. FOMM \cite{Siarohin2019fomm} applies first-order local affine transformations for motion flow transfer. Face-vid2vid \cite{wang2021facevid2vid} builds on this by introducing a learned 3D keypoint system for free-view head generation. StyleHEAT \cite{yin2022styleheat} integrates StyleGAN's latent features \cite{karras2019style} for motion and expression animation. Xue et al. \cite{Xue2023High} employ the Projected Normalized Coordinate Code (PNCC) \cite{zhu2017face} to enhance the preservation of facial details. They reconstruct the PNCC using the source identity parameters along with the target pose and expression parameters, which are estimated through 3D face reconstruction, effectively isolating the target identity. Ren et al. \cite{Ren2023HR} introduce HR-Net, a landmark-based approach that not only faithfully renders these expressions and postures across different identities but also enhances realism in face details and background consistency, supported by a novel loss function and extensive testing to achieve state-of-the-art results. Nevertheless, these methods are dependent on specific visual inputs, and the poses and expressions in the generated content are closely tied to these inputs, limiting their adaptability.

On the other hand, a more flexible solution is the audio-driven method, which creates talking heads from audio streams \cite{thies2020neural, wen2020photorealistic, Wen2020Photo, zhou2021pose, wang2022one, Eskimez2022speech}. This cutting-edge technique uses audio features like pitch and tone to animate still images, producing the appearance of speech. A major benefit of this method is its minimal reliance on visual data, enhancing its accessibility and adaptability. Current audio-driven technologies in video synthesis encompass several approaches. The first category involves end-to-end audio-to-image translation. For example, SyncNet \cite{Chung2016sync} introduces a dual-stream convolutional network that aligns audio inputs with corresponding mouth imagery. Wav2Lip \cite{prajwal2020lip} further refines this by improving lip-sync accuracy using a specialized lip-sync expert model. However, this direct audio-to-video mapping often limits control over detailed facial expressions and poses. To address this, another approach focuses on learning semantic audio representations and then rendering these into video. MakeItTalk \cite{zhou2020makeittalk} distinguishes between content and speaker attributes in the audio signal, using facial landmarks as intermediaries to produce video frames. Audio2Head \cite{wang2021audio2head} predicts dense motion fields from audio inputs and then uses a renderer for photo-realistic talking-head videos. Similarly, StyleHEAT \cite{yin2022styleheat} maps audio spectrograms to flow fields for animation. SadTalker \cite{zhang2023sadtalker} takes a unique approach by generating 3D motion coefficients from audio, which are then used to modulate a 3D-aware face renderer, facilitating realistic talking head synthesis. These methods aim to capture both lip synchronization and facial movements for more lifelike animations. Yi et al. \cite{Yi2023predict} address the one-to-many mapping problem of speech signals to talking head movements by proposing a novel two-step mapping strategy using a deep neural network model that predicts head motion from a speech signal and short video, outputting a high-fidelity talking face video with personalized head pose. However, a common limitation of these technologies is their reliance solely on audio clips from existing sources, overlooking the rich potential of audio-visual interactions and correlations for learning semantic representations. Recent advances in dynamic correlation learning \cite{chen2024dynamic} and heterogeneous semantic transfer \cite{chen2024heterogeneous} offer effective strategies for enhancing representation learning and knowledge transfer, offering new insights into improving temporal audio-visual correlation embedding for audio-driven one-shot talking head animation.

Different from aforementioned methods, we propose to investigate temporal audio-visual correlations as additional guidance and supervision to help facilitate the lip-synchronized performance and meanwhile improve the quality of final generated image.

\section{Methodology}

\begin{figure*}[t]
\centering
\includegraphics[width=0.85\textwidth]{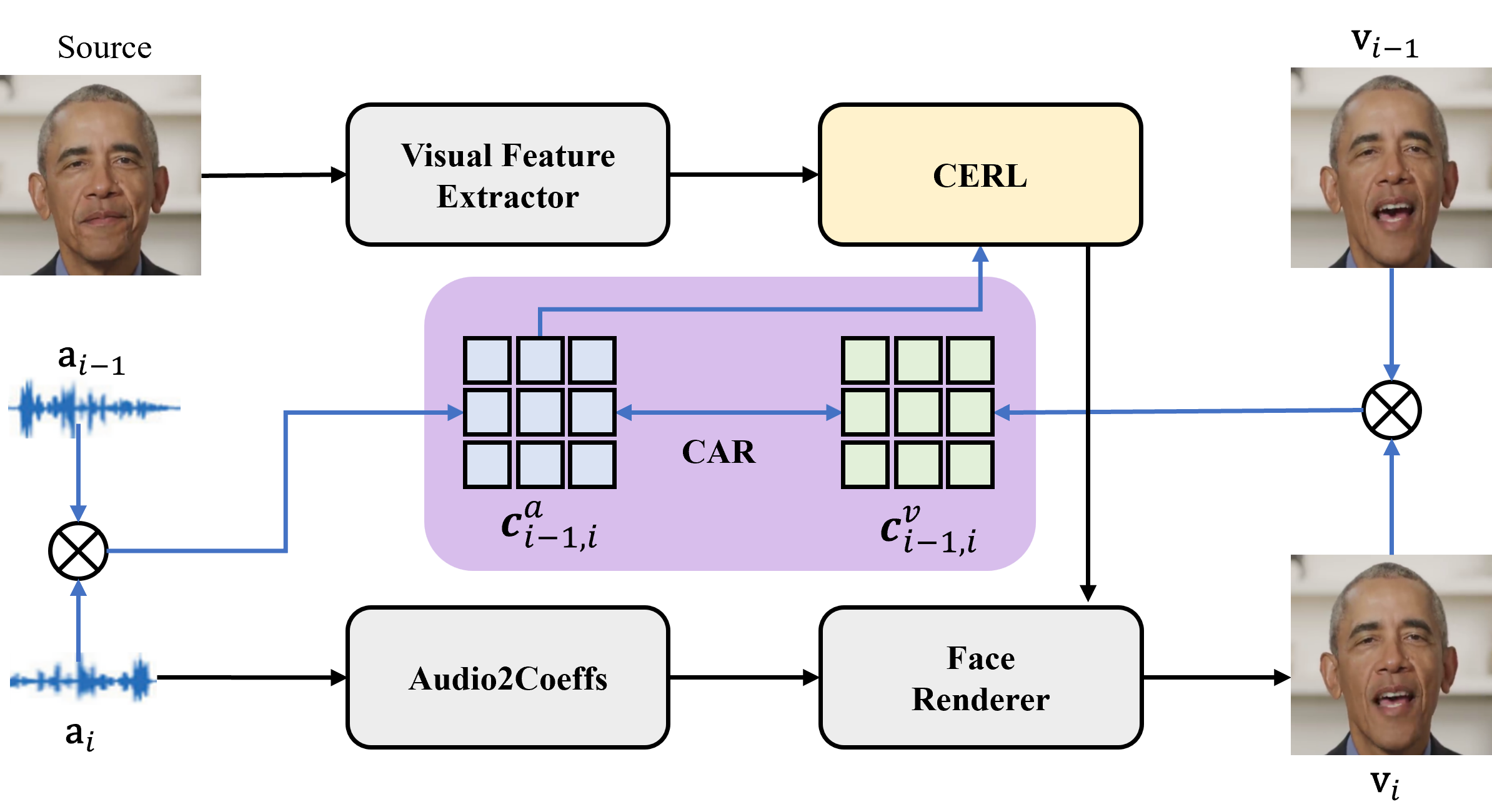}
\caption{An overall pipeline of the proposed Temporal Audio-Visual Correlation Embedding framework. Given a source image and the driven audio, it first extracts the image feature from the source image and predicts 3D coefficients from the audio. We then compute the temporal relationship between the current and previous audio clips, integrating this with the image feature for enhanced feature representation. Following this, the face renderer generates the final image from the image feature and the mapped 3D coefficients. Moreover, during training, the last real visual frame is used to calculate the temporal visual relationship with the generated image. The visual relationship is constrained to be similar to the audio relationship.} 
\label{fig:framework}
\end{figure*}

\subsection{Overview}
The Temporal Audio-Visual Correlation Embedding (TAVCE) framework first learns a temporal Audio-Visual correlation (TAVC) metric to align the temporal relationships of consecutive audio clips and corresponding video frames (Section \ref{sec:correlation_metric}). Then, we incorporate the temporal audio relationship to facilitate learning more representative feature representation since the temporal audio relationship contains information about the generated image (Section \ref{sec:enhancement}). Besides, the metric can be also used as an additional supervision signal to guide the generating of the final image (Section \ref{sec:regularization}). As shown in Figure \ref{fig:framework}, the temporal relationship $\mathbf{c}^v_{i-1,i}$ between the generated video clips $\mathbf{v}_{i-1}$ and $\mathbf{v}_{i}$ is expected to align with the temporal relationship $\mathbf{c}^a_{i-1,i}$ between given audio clips $\mathbf{a}_{i-1}$ and $\mathbf{a}_{i}$. When generating $\mathbf{v}_{i}$, we can use the temporal audio relationship $\mathbf{c}^a_{i-1,i}$ to help enhance feature representation and use the pre-trained TAVC metric to constrain the consistency between $\mathbf{c}^v_{i-1,i}$ and $\mathbf{c}^a_{i-1,i}$. The overall training procedure is presented in Algorithm \ref{alg:framework}.

\begin{algorithm}[h]
\caption{The training procedure of the proposed TAVCE framework.}
\label{alg:framework}
\begin{algorithmic}[1]
\STATE \textbf{Stage 1: Pre-train TAVC Metric}
\FOR{each sample $s$ in dataset}
    \FOR{$i = 1$ to $\text{len}(s)-1$}
        \STATE $a_i, a_{i-1} \gets s.\text{audios}[i], s.\text{audios}[i-1]$
        \STATE $v_i, v_{i-1} \gets s.\text{frames}[i], s.\text{frames}[i-1]$
        \STATE $c^a_{i,i-1} \gets \text{Cov}(E_a(a_{i-1}), E_a(a_i))$ \COMMENT{Eq. 2}
        \STATE $c^v_{i,i-1} \gets \text{Cov}(E_v(v_{i-1}), E_v(v_i))$ \COMMENT{Eq. 2}
        \STATE Minimize $\mathcal{L}_{\text{tavc}}$ \COMMENT{Eq. 4}
    \ENDFOR
\ENDFOR
\STATE
\STATE \textbf{Stage 2: Train Generation Framework}
\FOR{each sample $s$ in dataset}
    \FOR{$i = 1$ to $\text{len}(s)-1$}
        \STATE $a_i, a_{i-1} \gets s.\text{audios}[i], s.\text{audios}[i-1]$
        \STATE $j \gets \text{random\_int}(0, \text{len}(s)-1), j \neq i$
        \STATE $v_0, v_{i-1} \gets s.\text{frames}[j], s.\text{frames}[i-1]$
        \STATE $c^a_{i,i-1} \gets \text{Cov}(E_a(a_{i-1}), E_a(a_i))$ \COMMENT{Eq. 2}
        \STATE $f \gets E_f(v_0)$ \COMMENT{Extract visual feature}
        \STATE $g \gets \text{CERL}(f, c^a_{i,i-1})$ \COMMENT{Enhance feature, Eq. 5}
        \STATE $v_i \gets G(g, a_i)$ \COMMENT{Generate target frame}
        \STATE $c^{v'}_{i,i-1} \gets \text{Cov}(E_v(v_{i-1}), E_v(v_i))$ \COMMENT{Eq. 2}
        \STATE Minimize $\mathcal{L}_{\text{render}}$ \COMMENT{Original loss}
        \STATE Minimize $\mathcal{L}_{\text{reg}}$ \COMMENT{Additional supervision, Eq. 6}
    \ENDFOR
\ENDFOR
\end{algorithmic}
\end{algorithm}



\subsection{Temporal Audio-Visual Correlation Metric}
\label{sec:correlation_metric}
The TAVC metric aims to assign a high value to the correlation between the temporal relationships of consecutive audio clips and corresponding video frames and assign a small value otherwise. To this end, we first use the audio and visual encoders to extract audio and visual features. Then, we introduce a covariance matrix to compute the relationships between two audio clips or two image frames. We define a triple loss to obtain the above objective. 

Formally, given two audio clips $\{\mathbf{a}_{i}, \mathbf{a}_{j}\}$, we first  extract their features via an audio encoder $E^a(\cdot)$ to obtain their feature vector, denoted as 
\begin{align}
\begin{split}
\mathbf{f}^{a}_{i} & = E^a(\mathbf{a}_{i}) \\
\mathbf{f}^{a}_{j} & = E^a(\mathbf{a}_{j})
\label{eq:feature-dis}
\end{split}
\end{align}
The audio encoder is implemented by Audio2Coeff \cite{zhang2023sadtalker} followed by a fully-connected layer to obtain a 128-dimension vector. Audio2Coeff is described in detail in section \ref{sec:impl_details}. Then, we can compute their covariance matrix
\begin{equation}
    \mathbf{c}^a_{i,j}=\text{Cov}(\mathbf{f}^a_{i}, \mathbf{f}^a_{j})=\mathbb{E}[\mathbf{f}^a_{i}\mathbf{f}^a_{j}]-\mathbb{E}[\mathbf{f}^a_{i}]\mathbb{E}[\mathbf{f}^a_{j}]
    \label{eq:correlation}
\end{equation}
Covariance matrix \cite{fisher1936use} is a measure that indicates the extent to which two vectors change together, and it is widely used in statistical modeling for understanding the relationships of two given vectors. Here, we use it to measure the temporal audio relationships. Similarly, given two video frames $\{\mathbf{v}_{i}, \mathbf{v}_{j}\}$, we use a visual encoder to extract their feature vectors $\mathbf{f}^{v}_{i}, \mathbf{f}^{v}_{j}$ and also compute the covariance matrix $\mathbf{c}^v_{i,j}$ to denote their relationships. Here, the visual encoder is implemented by the ResNet18 \cite{he2016resnet} followed by a fully-connected layer to obtain a 128-dimension vector. Obviously, the temporal audio and visual relationships $\mathbf{c}^a_{i,j}$ and $\mathbf{c}^v_{i,j}$ are with dimension of $128\times 128$.

To learn the metric, we need to construct a dataset that covers positive and negative samples. Here, it is assigned to positive if giving two consecutive audio clips $\{\mathbf{a}_{i-1}, \mathbf{a}_{i}\}$ and the corresponding video frames $\{\mathbf{v}_{i-1}, \mathbf{v}_{i}\}$. The temporal audio and visual relationships, i.e., $\mathbf{c}^a_{i-1,i}$ and  $\mathbf{c}^v_{i-1,i}$ is a positive audio-visual correlation sample. In contrast, it is assigned to negative if giving two non-consecutive video clips $\{\mathbf{v}_{i-1}, \mathbf{v}_{j}\}$, in which $j$ does not belong to the interval $[i-\tau,i+\tau]$. Here, $\tau$ is a small integer value. The temporal audio and visual relationships, i.e., $\mathbf{c}^a_{i-1,i}$ and  $\mathbf{c}^v_{i-1,j}$ is negative audio-visual correlation sample. To maximize the similarity of positive samples while  minimizing that of the negative samples, the objective can be defined as
\begin{align}
\begin{split}
\ell_i&=[1-<\mathbf{c}^a_{i-1, i},\mathbf{c}^v_{i-1, i}>] + [1+<\mathbf{c}^a_{i-1, i},\mathbf{c}^v_{i-1, j}>] \\
&j\notin[i-\tau,i+\tau]
\label{eq:feature-dis2}
\end{split}
\end{align}
Here, where $<x, y>$ denotes the cosine similarity between $x$ and $y$. We compute the summation over all selected audio clips to obtain the final objective function, formulated as 
\begin{equation}
    \mathcal{L}_{tavc}=\sum_{i}\ell_i
    \label{eq:objective}
\end{equation}

When the TAVC metric is completely trained, it is expected the temporal audio relationship is aligned with the corresponding temporal visual relationship, and it can be treated as additional information to enhance feature representation learning and regularize final generation. We introduce a correlation-embedded representation learning (CERL) module to learn more representative feature representation and a correlation-aware regularization (CAR) loss to regularize the final generation process. In the following, we introduce these two modules in detail.

\begin{figure}[!t]
\centering
\includegraphics[width=0.48\textwidth]{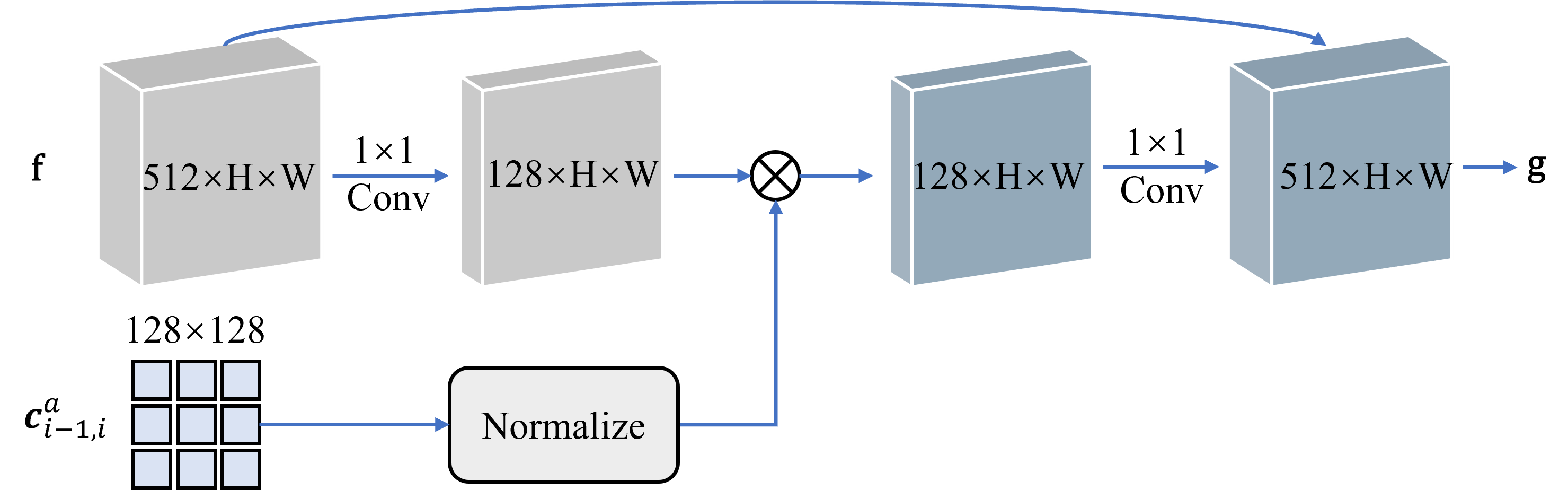}
\caption{Illustration of the correlation-embedded representation learning.} 
\label{fig:channel_attention}
\end{figure}

\subsection{Correlation-Embedded Representation Learning}\label{sec:enhancement}
Obviously, the temporal audio relationship $\mathbf{c}^a_{i-1, i}$ contains rich information to help generate the current video frame $\mathbf{v}_i$, and thus we use a correlation-embedded representation learning (CERL) module to integrate the temporal audio relationship to facilitate learning more representative feature representation that may capture more facial details. It is implemented via a simple yet effective channel attention mechanism. 

Given the source image $\textbf{v}$, we use a visual feature extractor implemented by ResNet18 \cite{he2016resnet} to extract its feature vector $\mathbf{f}\in\mathbb{R}^{512\times H\times W}$, where $H,W$ are the height and width of the feature map. When predicting the $i$-th frame, we use the temporal audio relationship $\mathbf{c}^a_{i-1, i}\in\mathbb{R}^{128\times128}$ computed by Equation \ref{eq:correlation}. Then, instead of directly fusing the image feature with the audio features, we introduce a channel attention mechanism to integrate the temporal audio relationship $\mathbf{c}^a_{i-1,i}$ to guide learning more powerful feature representation, as exhibited in Figure \ref{fig:channel_attention}. We first apply L2 normalization to normalize $\mathbf{c}^a_{i-1, i}$, and thus standardize its scale. Meanwhile, we use a $1\times 1$ convolution operation to reduce the channel dimension of the $\mathbf{f}$ to 128, followed by a matrix multiplication operation to obtain correlation-embedded representation. Finally, we apply a $1\times 1$ convolution operation to increase the channel dimension to the original 512 and combine the original image features to better preserve the identification information. The process can be expressed as 
\begin{equation}
    \mathbf{g}=\text{Conv}_1(\mathbf{c}^a_{i-1,i}\times \text{Conv}_2(\mathbf{f}))+\mathbf{f}
\end{equation}
where $\mathbf{g}$ is the enhanced feature and $\text{Conv}_1$ and $\text{Conv}_2$ are the two $1\times 1$ convolution operations.

\subsection{Correlation-Aware Regularization}\label{sec:regularization}
To improve the temporal coherence of the generated video frames, we introduce a Correlation-Aware Regularization (CAR) loss. This regularization is based on the observation that the temporal relationships in the audio domain should align with those in the visual domain once the TAVC metric is trained. By leveraging this correlation, we can guide the generation process to maintain consistency across frames.

During training, given a video frame $\mathbf{v}_{i-1}$ and its corresponding audio clips $\mathbf{a}_{i-1}$ and $\mathbf{a}_{i}$, our objective is to generate the target visual frame $\mathbf{v}_{i}$. We can compute both temporal audio relationship $\mathbf{c}^a_{i-1, i}$ and temporal visual relationship $\mathbf{c}^v_{i-1, i}$ via Equation \ref{eq:correlation}. It is obvious that $\textbf{c}^a_{i-1, i}$ and $\textbf{c}^v_{i-1, i}$ are positive audio-visual correlation samples, and their similarity should be high.

To explicitly enforce this consistency, we introduce the following regularization objective:
\begin{equation}
    \mathcal{L}_{reg} = \frac{1}{N}\sum_{i=1}^N [1-<\mathbf{c}^a_{i-1, i},\mathbf{c}^v_{i-1, i}>]
    \label{eq:supervision}
\end{equation}
This loss term encourages the generated visual content to exhibit temporal transitions that are in close alignment with the audio transitions. By integrating this additional regularization, we leverage the pre-trained TAVC metric as an extra supervisory signal, ensuring that the model not only learns to generate plausible individual frames but also maintains consistent temporal dynamics as dictated by the audio.

\subsection{Implementation Details}\label{sec:impl_details}
In this part, we introduce the details of the other modules in our framework, as well as the training details. Audio2Coeff \cite{zhang2023sadtalker} is adept at deriving precise facial expression and head pose coefficients from any given audio input. It is divided into two specialized sub-models, each is used to predict one of these coefficient sets. For the facial expression component, the network's design incorporates a ResNet-based \cite{he2016resnet} audio encoder, followed by a linear layer. When a continuous audio clip of $t$ frames is input, this data is first converted to a mel-spectrum, and then it is fed into the model together with the reference facial expression coefficients. The output is facial expression coefficients of $t$ frames that match the audio content. On the other hand, head pose prediction utilizes a VAE \cite{kingma2013vae} model with an encoder-decoder structure. Both the encoder and the decoder are constructed using two MLP layers. Upon receiving a continuous audio clip of $t$ frames, the model initially extracts an audio embedding through an audio encoder. This embedding data is then fed into the VAE, generating $t$ frames of head pose residuals. Finally, these residuals are added to the initial reference pose, resulting in the predicted head pose.

Face Renderer utilizes the coefficients determined in the prior step to transform the source face into the target face. We employ the cutting-edge image animation technique, face-vid2vid \cite{wang2021facevid2vid}, as our foundational model. This model is composed of a visual feature extractor, a canonical keypoint estimator, and a generator. Initially, the visual feature extractor extracts features from a source image. Subsequently, we utilize a mapping network composed of four 1D convolution layers to predict motion coefficients based on 3DMM \cite{blanz1999morphable} coefficients. The keypoint estimator then determines the canonical keypoints from the source image and forms the deformation matrix to warp the image feature. Finally, the generator produces the target face image. To facilitate coherent talking head generation, we integrate the correlations to help enhance feature representation and regularize final generation.

For training details, we adopt a two-stage approach to train our model. In the first stage, the audio-visual correlation module is trained using Equation \ref{eq:objective}. Following this, we incorporate the proposed temporal audio-visual correlation embedding framework into the Face Renderer. We then pretrain the Face Renderer according to the settings used in face-vid2vid \cite{wang2021facevid2vid}, employing a video-driven approach. In the second stage, we finetune the MappingNet with Face Renderer to optimize it for audio-driven talking head generation. The training iterations for the audio-visual correlation module and MappingNet are 40k and 12.5k, respectively. We set learning rates at $1e^{-4}$ for the former and $2e^{-4}$ for the latter.

\begin{table*}[htbp]
\centering
\caption{Comparison of FID, CSIM, LSE-D, and LSE-C metics of state-of-the-art methods and our TAVCE framwork on the HDTF, LRW, VoxCeleb1, and VoxCeleb2 datasets.}
\begin{tabular}{c|cc|cc|cc|cc}
\toprule
\multirow{3}{*}{Methods} &  \multicolumn{4}{c|}{HDTF} & \multicolumn{4}{c}{LRW} \\
\cline{2-9}
& \multicolumn{2}{c|}{Video Quality} & \multicolumn{2}{c|}{Lip Synchronization} & \multicolumn{2}{c|}{Video Quality}  & \multicolumn{2}{c}{Lip Synchronization } \\
\cline{2-9}
& FID$\downarrow$ & CSIM$\uparrow$ & LSE-D$\downarrow$ & LSE-C$\uparrow$ & FID$\downarrow$ & CSIM$\uparrow$ & LSE-D$\downarrow$ & LSE-C$\uparrow$ \\
\hline
Audio2Head & 14.234 & 0.820 & \textbf{7.405} & \textbf{7.535} & 11.708 & 0.736 & 8.048 & 5.734\\
MakeItTalk & 17.085 & 0.834 & 9.953 & 5.105 & 9.609 & 0.824 & 9.968 & 3.302\\
StyleHEAT & 15.691 & 0.757 & 7.949 & 6.811 & 20.698 & 0.442 & 7.710 & 5.893\\
SadTalker & 8.764 & 0.856 & 7.690 & 7.376 & 4.713 & 0.863 & 7.557 & 6.135\\
TAVCE (Ours) & \textbf{7.742} & \textbf{0.859} & 7.562 & 7.399 & \textbf{4.102} & \textbf{0.867} & \textbf{7.292} & \textbf{6.377} \\
\hline
\hline
& \multicolumn{4}{c|}{VoxCeleb1} & \multicolumn{4}{c}{VoxCeleb2} \\
\cline{2-9}
Audio2Head & 42.214 & 0.666 & 8.421 & 5.538 & 36.504 & 0.537 & 8.673 & 5.752 \\
MakeItTalk & 31.405 & 0.752 & 9.680 & 4.175 & 19.337 & 0.711 & 10.606 & 3.915\\
StyleHEAT & 54.887 & 0.442 & 8.570 & 5.241 & 48.128 & 0.386 & 9.623 & 4.379\\
SadTalker & 24.344 & \textbf{0.785} & 8.152 & 5.919 & 14.936 & 0.731 & 8.163 & 6.194\\
TAVCE (Ours) & \textbf{22.732} & 0.784 & \textbf{8.086} & \textbf{5.971} & \textbf{14.198} & \textbf{0.736} & \textbf{7.927} & \textbf{6.463}\\
\bottomrule
\end{tabular}
\label{table:quantitative_comparison}
\end{table*}

\section{Experiments}

\subsection{Experimental Settings}
\subsubsection{Datasets} 
The VoxCeleb2 \cite{Chung18bvox2} dataset comprises over 1 million utterances from 6,000 distinct speakers. For pre-training, we randomly chose 280,000 videos to pretrain both the audio-visual correlation module and the FaceRender module. During the fine-tuning phase, we selected 200 speakers at random, yielding 8,385 videos. To evaluate the performance of ADOS-THA, we adopt multiple datasets including the whole HDTF \cite{zhang2021hdtf} dataset and the test parts of the LRW, VoxCeleb1, and VoxCeleb2 datasets. The HDTF dataset contains 362 speakers. After removing some corrupted videos, there are a total of 373 videos. The LRW \cite{Chung16lrw} test set comprises 500 distinct words, for each of which we randomly select 3 videos, resulting in a total of 1500 test videos. The test sets of VoxCeleb1 \cite{Nagrani17vox1} and VoxCeleb2 \cite{Chung18bvox2} contain 50 and 118 speakers, respectively. Similarly, we randomly chose 3 videos for each speaker to construct our test dataset. 

\subsubsection{Evaluation Protocol}
In this work, we evaluate the performance using the following metrics: 1) Frechet Inception Distance (FID) \cite{heusel2017gans}: This metric obtains feature vectors from both generated and real videos using a pretrained state-of-the-art convolutional network \cite{szegedy2016inception}. The difference in these feature vectors is then computed to assess the realism of the generated video. A lower FID value signifies superior realism. 2) Cosine Similarity (CSIM): This metric derives features from both videos using a cutting-edge face recognition network \cite{deng2019arcface}. It then calculates the differences to measure the identity similarity between the generated and real videos. A higher CSIM value indicates a greater degree of similarity. 3) Lip Sync Error-Distance (LSE-D) and Lip Sync Error-Confidence (LSE-C) \cite{prajwal2020lip}: These metrics compute the distance and confidence, respectively, between lip and audio representations using a pre-trained model \cite{Chung2016sync}. They are utilized to evaluate the synchronization accuracy between lip movements and audio.

\subsection{Comparisons with State-Of-The-Art methods}
We conduct a comprehensive comparison of various state-of-the-art methods in the realm of talking head video generation. The selection of baseline methods is based on their prominence in the field, diversity in methodological approaches, and availability of publicly released pre-trained models, ensuring fair and reproducible evaluation. Specifically, we include methods that represent different paradigms in talking head generation: Audio2Head \cite{wang2021audio2head} generates dense motion fields from audio inputs, followed by a rendering process to produce photo-realistic talking-head videos. MakeItTalk \cite{zhou2020makeittalk} differentiates between content and speaker attributes within the audio signal to enhance animation realism. StyleHEAT \cite{yin2022styleheat} utilizes audio spectrograms to generate flow fields, driving facial animation effectively. SadTalker \cite{zhang2023sadtalker} derives 3D motion coefficients from audio and employs a 3D-aware face renderer to create lifelike talking-head videos.

\subsubsection{Quantitative Comparisons.}
We present the performance comparisons on the 4 test datasets in Table \ref{table:quantitative_comparison}. Our proposed framework achieves superior video quality and lip synchronization. We first analyze the performance on the widely used HDTF dataset. Compared to the current top-performing approach, SadTalker, our framework shows significant improvements in video quality. It decreases the FID from 8.764 to 7.742 and increases the CSIM from 0.856 to 0.859. In terms of lip synchronization, our framework decreases LSE-D from 7.690 to 7.562 and increases LSE-C from 7.376 to 7.399, demonstrating better alignment between audio and visual elements. The audio2head method pays more attention to mouth shape but compromises the overall image quality. Although it achieves better LSE-D and LSE-C metrics, from video examples, our framework shows better lip synchronization, which can be verified by the user study we did. These notable advancements can be attributed to temporal audio-visual correlation's pivotal role in enhancing image feature representations and regularizing the training of the model.

LRW is a relatively simple dataset, with each audio segment containing only one word. our framework also shows significant advantages in both video quality and lip synchronization on this dataset. Specifically, our framework achieves a FID of 4.102 and a CSIM of 0.867, outperforming SadTalker by 13.0\% and 0.5\%, signifying improved visual quality. Furthermore, the LSE-D and LSE-C are measured at 7.292 and 6.377, with a relative reduction of 3.5\% and a relative increase of 3.9\% compared with SadTalker, suggesting a higher lip synchronization accuracy. 

VoxCeleb1 and VoxCeleb2 are audio-visual datasets featuring short clips of human speech from various speakers. The video clarity in these datasets is not as high as in HDTF, resulting in lower-quality output from associated methods. Despite this, our approach attains state-of-the-art performance. Notably, it records the best FID of 22.732 on the VoxCeleb1 dataset. The CSIM score is marginally lower at 0.784 compared to SadTalker's 0.785 but remains competitive. Moreover, our framework excels in lip synchronization, achieving LSE-D and LSE-C of 8.086 and 5.971, respectively. A similar trend can be observed on the VoxCeleb2 dataset. our framework achieves the best results in terms of video quality and lip synchronization. Specifically, it boasts a FID of 14.198, a CSIM score of 0.736, a LSE-D of 7.927, and a LSE-C of 6.463. Compared to SadTalker, these represent relative increases/decreases of 4.9\%, 0.7\%, 2.9\%, and 4.3\%, respectively.

\begin{figure*}[htbp]
\centering
\includegraphics[width=0.85\textwidth]{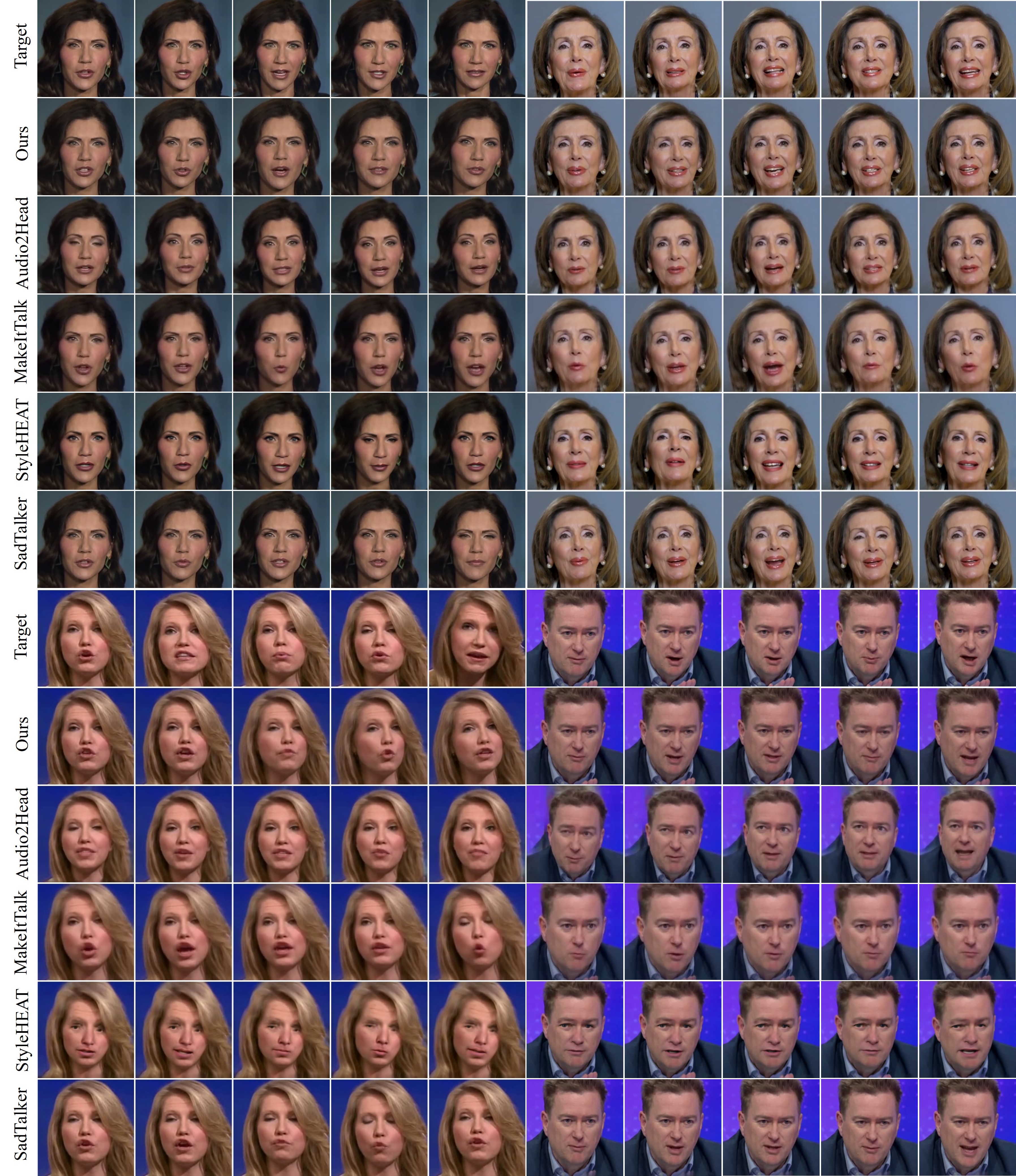}
\vspace{-10pt}
\caption{Qualitative comparisons of state-of-the-art methods and our TAVCE framwork for audio-driven one-shot talking head generation on the HDTF and LRW dataset. Our framework delivers high-quality generations in terms of lip synchronization and overall image quality.}
\label{fig:comparison_hdtf_lrw}
\end{figure*}

\subsubsection{Qualitative Comparisons.}
\begin{figure*}[htbp]
\centering
\includegraphics[width=1.0\textwidth]{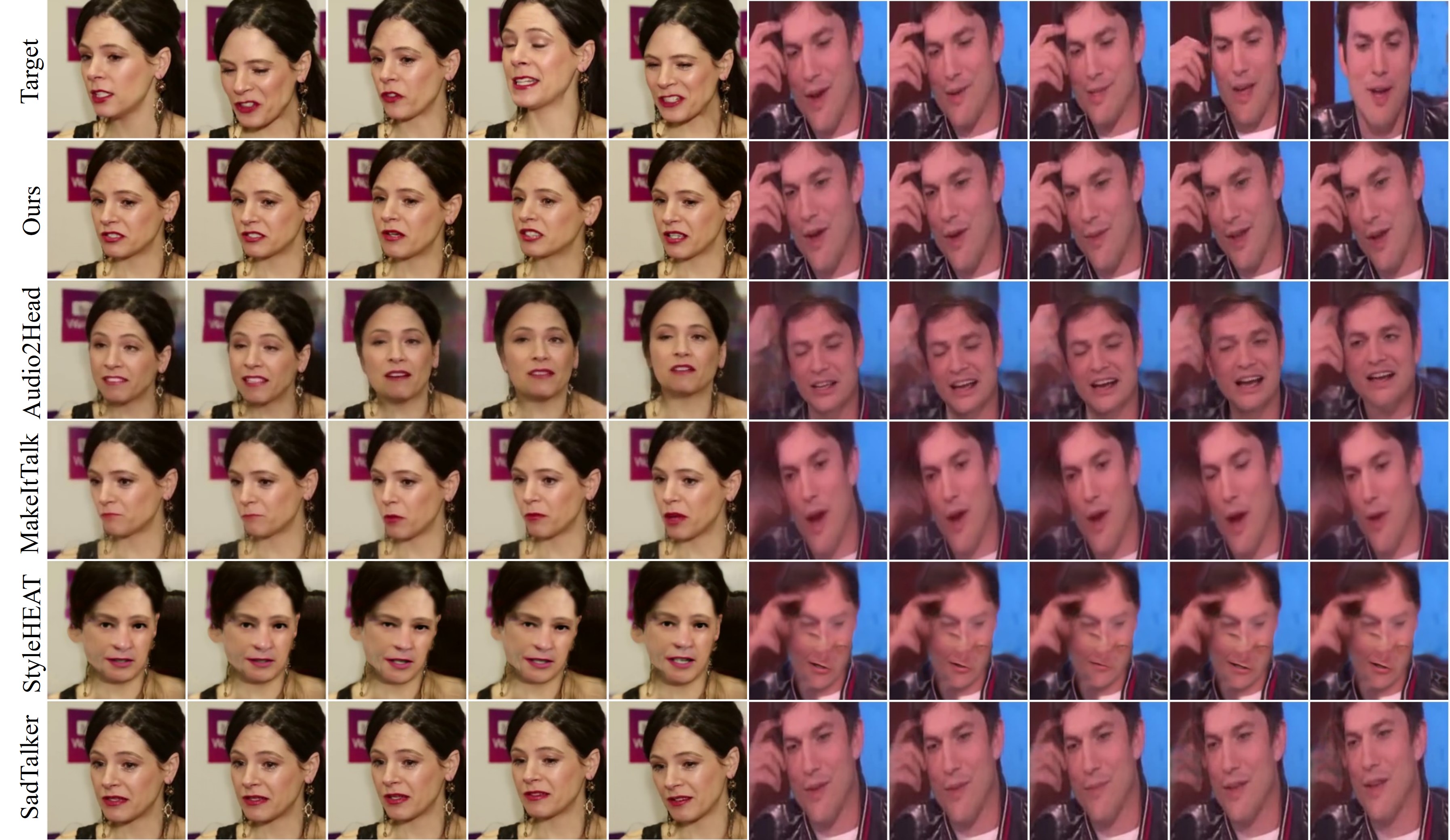}
\caption{Qualitative comparisons on the VoxCeleb1 and VoxCeleb2 dataset. Our framework achieves high-quality talking head animations, both in terms of lip synchronization and image quality.}
\label{fig:comparison_voxceleb1_2}
\end{figure*}

In this section, we present a comparative visualization of the results obtained from our framework alongside several leading-edge techniques, as illustrated in Figure \ref{fig:comparison_hdtf_lrw}. For reference, the top row displays target images that serve as benchmarks for both lip shape and identity. Similar to the quantitative metrics, we also analyze the qualitative comparisons from two aspects. 1) \textbf{Video quality.} Audio2Head and StyleHEAT experience issues with identity distortion. For instance, in the bottom left example, the eye area appears unnatural, and there is a noticeable decrease in identity similarity compared to other methods. MakeItTalk produces videos with blurred facial features, and lacking detail. This is evident in the bottom left example, where areas around the eyes and mouth are blurry, and facial textures appear overly smooth. While both SadTalker and our framework, utilizing face-vid2vid as a renderer, achieve superior overall video quality, SadTalker's outputs sometimes exhibit incoherence. As shown in the bottom right example, SadTalker's output shows a vague finger across different frames, whereas our framework consistently renders fingers. 2) \textbf{Lip synchronization.} Current methods for generating dynamic lip movements from audio input fall short in accuracy under certain conditions. A comparison with target images reveals that the mouth shapes produced by existing techniques sometimes differ significantly from actual mouth shapes. For instance, in the second frame of the top left example, the desired mouth shape is open, yet Audio2Head, StyleHEAT, and SadTalker yield a closed mouth. Similarly, the third frame requires an open mouth, but MakeItTalk incorrectly presents a closed mouth. In contrast, our framework achieves higher lip synchronization accuracy by learning the temporal audio-visual correlation metric to supervise the generations of the model. More qualitative comparisons on the VoxCeleb1 and VoxCeleb2 dataset are presented in Figure \ref{fig:comparison_voxceleb1_2}. 

\subsubsection{User Study}
We conduct online user studies to perform comparative analyses of various methods. This study is divided into two aspects, corresponding to the metrics previously discussed: video quality and lip synchronization accuracy. For each dataset, we create 20 test videos showcasing a range of ages, genders, and facial expressions. A panel of 20 participants is recruited, with each member tasked with evaluating both metrics for each video. As shown in Table \ref{table:user_study}, our framework outperforms other state-of-the-art techniques in both aspects. For instance, on the HDTF dataset, our framework achieves a 38.2\% rating in video quality and a 42.8\% rating in lip synchronization, surpassing SadTalker by 13.2\% and 21.3\%, respectively. Similar superior performance is observed on the LRW, VoxCeleb1, and VoxCeleb2 datasets. \emph{All test videos utilized in the user study are included at \url{https://zh-xu410.github.io/TAVCE-Suppl/}}.

\begin{table}[h]
\centering
\caption{User studies on video quality and lip synchronization on the HDTF, LRW, VoxCeleb1, and VoxCeleb2 datasets.}
\begin{tabular}{c|cc|cc}
\toprule
\multirow{2}{*}{Methods} & \multicolumn{2}{c|}{HDTF} & \multicolumn{2}{c}{LRW}  \\
\cline{2-5}
& Quality & Lip Sync & Quality & Lip Sync \\
\hline
Audio2Head & 18.0\% & 19.0\% & 15.0\% & 15.0\% \\
MakeItTalk & 4.8\% & 4.5\% & 7.5\% & 6.2\% \\
StyleHEAT & 14.0\% & 12.2\% & 6.2\% & 6.2\% \\
SadTalker & 25.0\% & 21.5\% & 30.0\% & 28.8\% \\
TAVCE (Ours) & \textbf{38.2\%} & \textbf{42.8\%} & \textbf{41.2\%} & \textbf{43.8\%} \\
\hline
\hline
& \multicolumn{2}{c|}{VoxCeleb1} & \multicolumn{2}{c}{VoxCeleb2}  \\
\cline{2-5}
Audio2Head & 13.8\% & 12.5\% & 7.8\% & 10.0\% \\
MakeItTalk & 17.5\% & 17.5\% & 6.7\% & 6.7\% \\
StyleHEAT & 5.0\% & 7.5\% & 10.0\% & 8.9\% \\
SadTalker & 30.0\% & 28.8\% & 33.3\% & 36.7\% \\
TAVCE (Ours) & \textbf{33.8\%} & \textbf{33.8\%} & \textbf{42.2\%} & \textbf{37.8\%} \\
\bottomrule
\end{tabular}
\label{table:user_study}
\end{table}

\begin{table}[h]
\centering
\caption{Complexity comparison of different methods.}
\begin{tabular}{ccc}
\toprule
Methods & FLOPs (G) & Params (M)  \\
\hline
Audio2Head & 227.1 & 144.7 \\
MakeItTalk & 31.9 & 72.9 \\
StyleHEAT & 115.7 & 368.1 \\
SadTalker & 623.5 & 175.4 \\
TAVCE (Ours) & 623.7 & 178.3 \\
\bottomrule
\end{tabular}
\label{table:complexity}
\end{table}

\subsubsection{Complexity Analysis}

To ensure a fair comparison, we calculate the FLOPs based on the computational cost of generating a single frame, with both the input and output image resolutions uniformly set to 256×256 across all methods. This standardization allows for an equitable evaluation of computational complexity across different approaches. The results are summarized in Table \ref{table:complexity}.

For time complexity, we observe that Audio2Head, MakeItTalk, and StyleHEAT have relatively low FLOPs but produce less realistic results, with FID scores of 14.2, 17.1, and 15.7 on the HDTF dataset. In contrast, both our framework (TAVCE) and SadTalker demand higher computational costs due to their advanced capabilities. However, this increased cost is justified by substantial improvements in generation quality, with FID scores of 7.7 and 8.8, respectively. Moreover, compared to SadTalker, our framework incurs minimal overhead (0.03\%) while significantly enhancing performance, achieving a 12.5\% reduction in FID.

Regarding space complexity, the number of parameters reflects the memory demands of each model. Our model contains 178.3M parameters, slightly more than SadTalker’s 175.4M, yet significantly fewer than StyleHEAT’s 368.1M. Despite this modest increase, the performance gains outweigh the additional parameters, further underscoring the efficiency of our enhancements. This balance between computational efficiency and model performance highlights the effectiveness of our method.

\subsection{Ablative Studies}

The above comparisons with state-of-the-art methods well demonstrate the effectiveness of the proposed TAVCE framework as a whole. In this part, we further delve into a detailed module to analyze their actual contributions. The results are evaluated on the HDTF dataset.

\subsubsection{Effectiveness of Audio-Visual Correlation Embedding}

\begin{table}[htbp]
\small
\centering
\setlength{\tabcolsep}{3pt}
\caption{Comparison of FID, CSIM, LSE-D, and LSE-C metics with and without the proposed modules, where CERL is the correlation-embedded representation learning and CAR is the correlation-aware regularization.}
\begin{tabular}{c|cc|cc}
\toprule
\multirow{2}{*}{Methods} & \multicolumn{2}{c|}{Video Quality}  & \multicolumn{2}{c}{Lip Synchronization } \\
\cline{2-5}
& FID$\downarrow$ & CSIM$\uparrow$ & LSE-D$\downarrow$ & LSE-C$\uparrow$ \\
\hline
Ours w/o CERL & 7.914 & 0.858 & 7.719 & 7.344 \\
Ours w/o CAR & 8.029 & 0.858 & 7.775 & 7.216 \\
Ours w/o CERL \& CAR & 8.263 & 0.857 & 7.882 & 7.137 \\
Ours & \textbf{7.742} & \textbf{0.859} & \textbf{7.562} & \textbf{7.399} \\
\bottomrule
\end{tabular}
\label{table:ablative}
\end{table}

Our proposed TAVCE method hinges on two critical components: correlation-aware regularization and correlation-aware representation. To ascertain the individual contribution of each, we perform a series of ablative experiments that selectively exclude either one or both of these elements. As shown in Table \ref{table:ablative}, compared with the baseline model without the two modules, our framework decreases the FID, LSE-D from 8.263, 7.882 to 7.742, 7.562, with the decrease of 6.3\%, 4.1\%; and increases the CSIM, LSE-C from 0.857, 7.137 to 0.859, 7.399, with the increase of 2.3\%, 3.7\%. It well demonstrates the effectiveness of our proposed TAVCE method.

We further analyze the similarity between audio and visual temporal relationships. We calculate the average cosine similarity for both positive (adjacent) and negative (non-adjacent) samples. The similarity for positive samples stands at 0.507, which is notably three times higher than the 0.158 score observed for negative samples. This finding underscores a strong correlation between the relationships of adjacent audio clips and visual frames, which can provide rich information for model guidance.

\begin{figure}[h]
    \centering
    \includegraphics[width=0.5\textwidth]{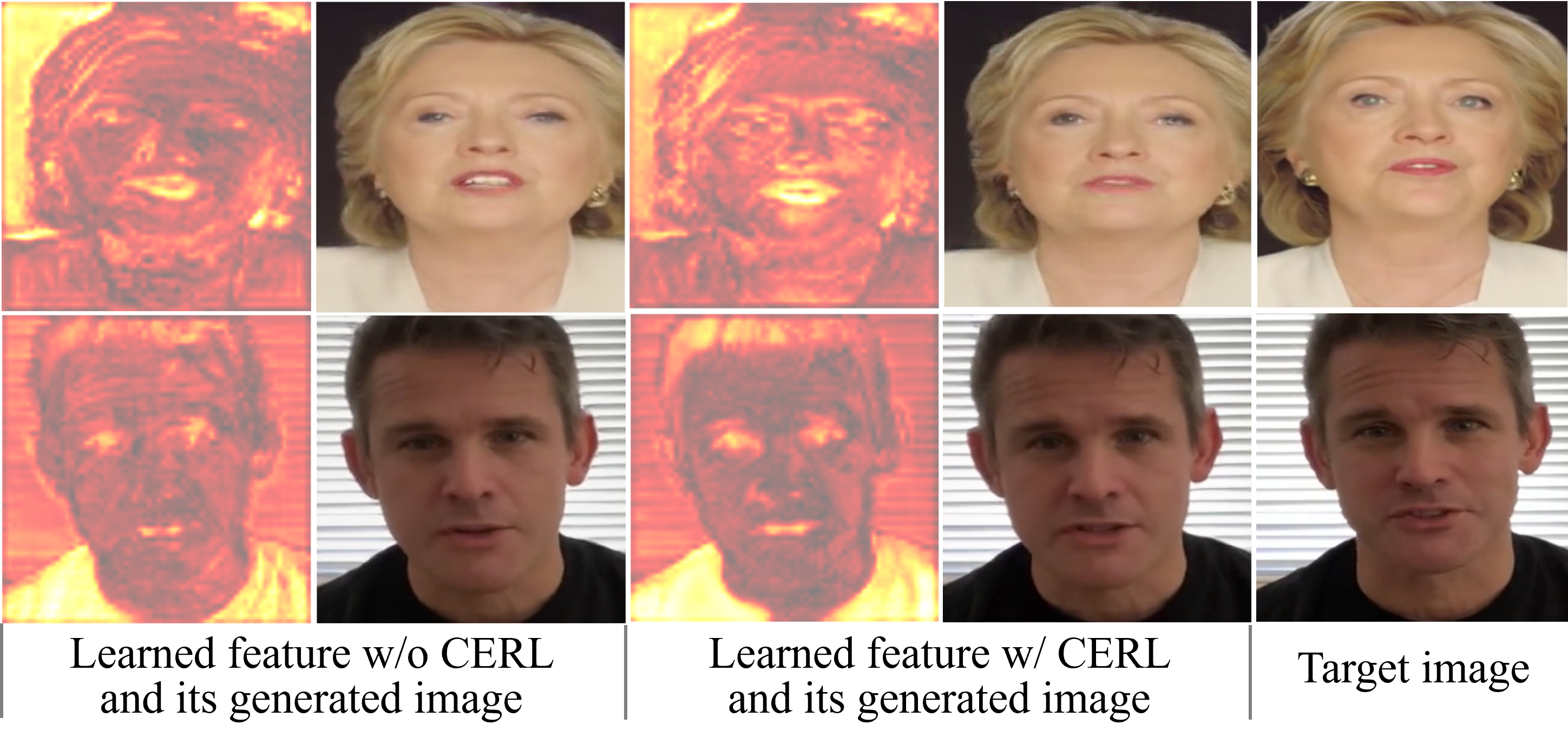}
    \caption{Comparison of learned features with and without CERL and their corresponding generated images. The learned features enhanced by CERL module emphasize more on the mouth area and thus keep more details to better keep the mouth animations.}
    \label{fig:feature}
\end{figure}
\vspace{-10pt}

\subsubsection{Analysis of Correlation-Embedded Representation Learning}
The Correlation-Embedded Representation Learning (CERL) module integrates temporal audio relationships with video frame features to leverage contextual information. To assess its impact, we conducted an experiment where it was omitted from our framework. As evidenced in Table \ref{table:ablative}, the absence of the CERL module results in a marked deterioration of performance: the FID increases from 7.742 to 7.914, and the LSE-D increases from 7.562 to 7.719. Additionally, the CSIM and LSE-C also exhibit a marginal decline. 

To have a more intuitive understanding of the effect of the CAE module, we present some examples of feature visualization in Figure \ref{fig:feature}. Upon comparing the feature maps with and without CERL, it's evident that the features produced by CERL show higher heat values at the mouth area. Consequently, the mouth shape in images generated from these enhanced features is more accurately depicted than in those derived from non-enhanced features. For example, in the first row, the model produces an image with an open mouth when using the non-enhanced feature, whereas the target is actually a nearly closed mouth. Conversely, the mouth shape generated from the enhanced features more closely resembles the target mouth shape.

\subsubsection{Analysis of Correlation-Aware Regularization}
The Correlation-Aware Regularization (CAR) module incorporates the temporal audio-visual correlation metric as an auxiliary objective to ensure that the generated frames are natural and coherent. It demands that the temporal relationship of the generated visual frame and the last real frame be consistent with the temporal relationship of corresponding adjacent audios. As shown in Table \ref{table:ablative}, removing the CAR module from our framework, the FID, LSE-D increases from 7.742, 7.562 to 8.029, 7.775, and the CSIM, LSE-C decreases from 0.859, 7.339 to 0.858, 7.216, respectively.

\section{Limitations}

\begin{figure}[h]
    \centering
    \includegraphics[width=0.48\textwidth]{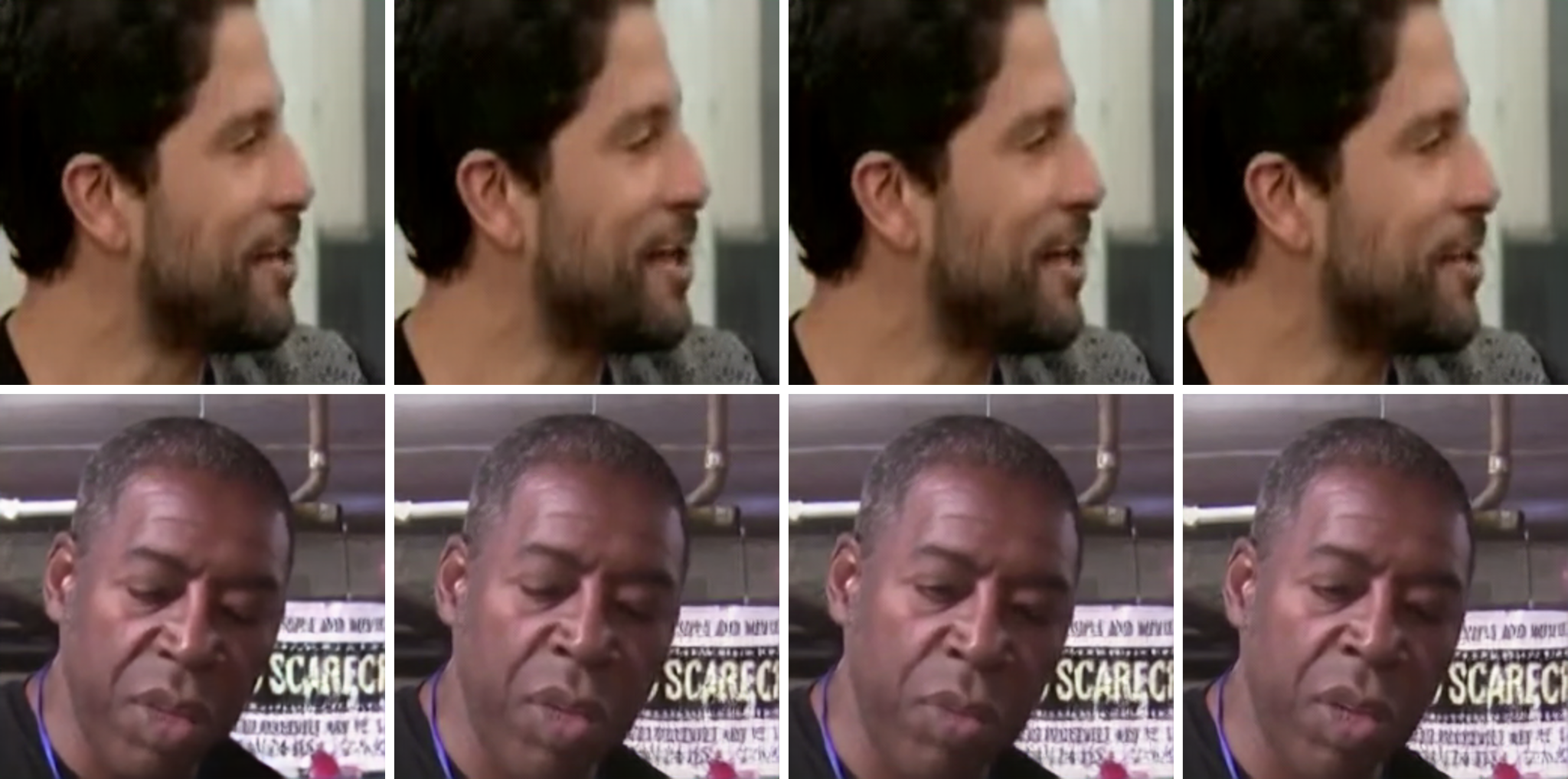}
    \caption{Examples of failure cases. The errors may arise due to challenging head poses and expressions.}
    \label{fig:failure_cases}
\end{figure}

While the framework demonstrates strong generalization across multiple datasets, certain limitations remain. It may still struggle in cases of extreme head movements and unseen expressions. Figure \ref{fig:failure_cases} presents failure cases of our framework. In the first row, a nearly 90-degree side head pose results in a blurry mouth, revealing limitations in handling extreme poses. The second row shows unrealistic facial expressions, with half-closed eyes and a blurry mouth, suggesting insufficient learning of expression diversity. Enhancing the model’s adaptability to such scenarios remains an open challenge.

To address this, future work will explore the usage of 3D Morphable Models (3DMM) to achieve a more precise disentanglement of mouth movements, head poses, and facial expressions. By independently modeling these components, we aim to better capture their temporal relationships and establish more accurate correlations between audio sequences and these disentangled representations. This approach will provide the model with finer-grained supervision, enhancing its ability to generate natural and coherent facial animations even in challenging conditions.

\section{Conclusion}
In this work, we propose a temporal audio-visual correlation embedding (TAVCE) framework to exploit the temporal audio-visual correlations to enhance feature representation and provide additional supervision to improve ADOS-THA performance. It first learns correlations between temporal audio relationships and temporal visual relationships. Subsequently, the temporal audio relationships are integrated with image features to learn more representative features, while the learned audio-visual correlation is used to regularize the generation. We conduct extensive experiments and carry out variant quantitative and qualitative comparisons as well as user studies to demonstrate the effectiveness.

\bibliographystyle{IEEEtran}
\bibliography{reference}

\begin{IEEEbiography}
[{\includegraphics[width=1in,height=1.25in,clip,keepaspectratio]{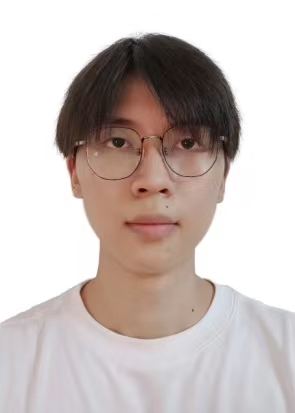}}]{Zhihua Xu} received the B.E. degree from the School of Computer Science and Technology, Guangdong University of Technology, Guangzhou, China. He is currently working toward the M.S. degree at the School of Information Engineering, also at Guangdong University of Technology. His current research interests include computer vision and deep learning.
\end{IEEEbiography}

\begin{IEEEbiography}[{\includegraphics[width=1in,height=1.25in,clip,keepaspectratio]{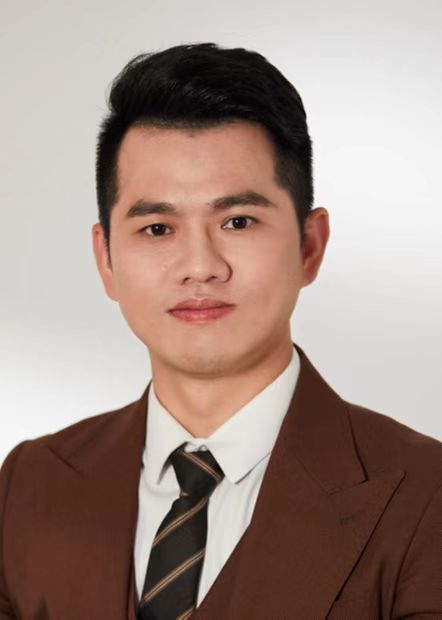}}]{Tianshui Chen} received a Ph.D. degree in computer science at the School of Data and Computer Science Sun Yat-sen University, Guangzhou, China, in 2018. Prior to earning his Ph.D, he received a B.E. degree from the School of Information and Science Technology in 2013. He is currently an associated professor in the Guangdong University of Technology. His current research interests include computer vision and machine learning. He has authored and coauthored more than 40 papers published in top-tier academic journals and conferences, including T-PAMI, T-NNLS, T-IP, T-MM, CVPR, ICCV, AAAI, IJCAI, ACM MM, etc. He has served as a reviewer for numerous academic journals and conferences. He was the recipient of the Best Paper Diamond Award at IEEE ICME 2017. \end{IEEEbiography}

\begin{IEEEbiography}[{\includegraphics[width=1in,height=1.25in,clip,keepaspectratio]{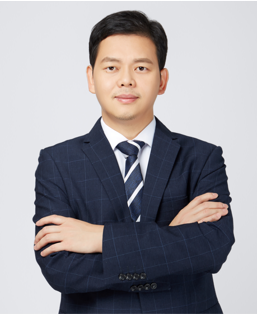}}]{Zhijing Yang} received the B.S and Ph.D. degrees from the Mathematics and Computing Science, Sun Yat-sen University, Guangzhou China, in 2003 and 2008, respectively. He was a Visiting Research Scholar in the School of Computing, Informatics and Media, University of Bradford, U.K, between July-Dec, 2009, and a Research Fellow in the School of Engineering, University of Lincoln, U.K, between Jan. 2011 to Jan. 2013. He is currently a Professor and Vice Dean at the School of Information Engineering, Guangdong University of Technology, China. He has published over 80 peer-review journal and conference papers, including IEEE T-CSVT, T-MM, T-GRS, PR, etc. His research interests include machine learning and pattern recognition.
\end{IEEEbiography}

\begin{IEEEbiography}[{\includegraphics[width=1in,height=1.25in,clip,keepaspectratio]{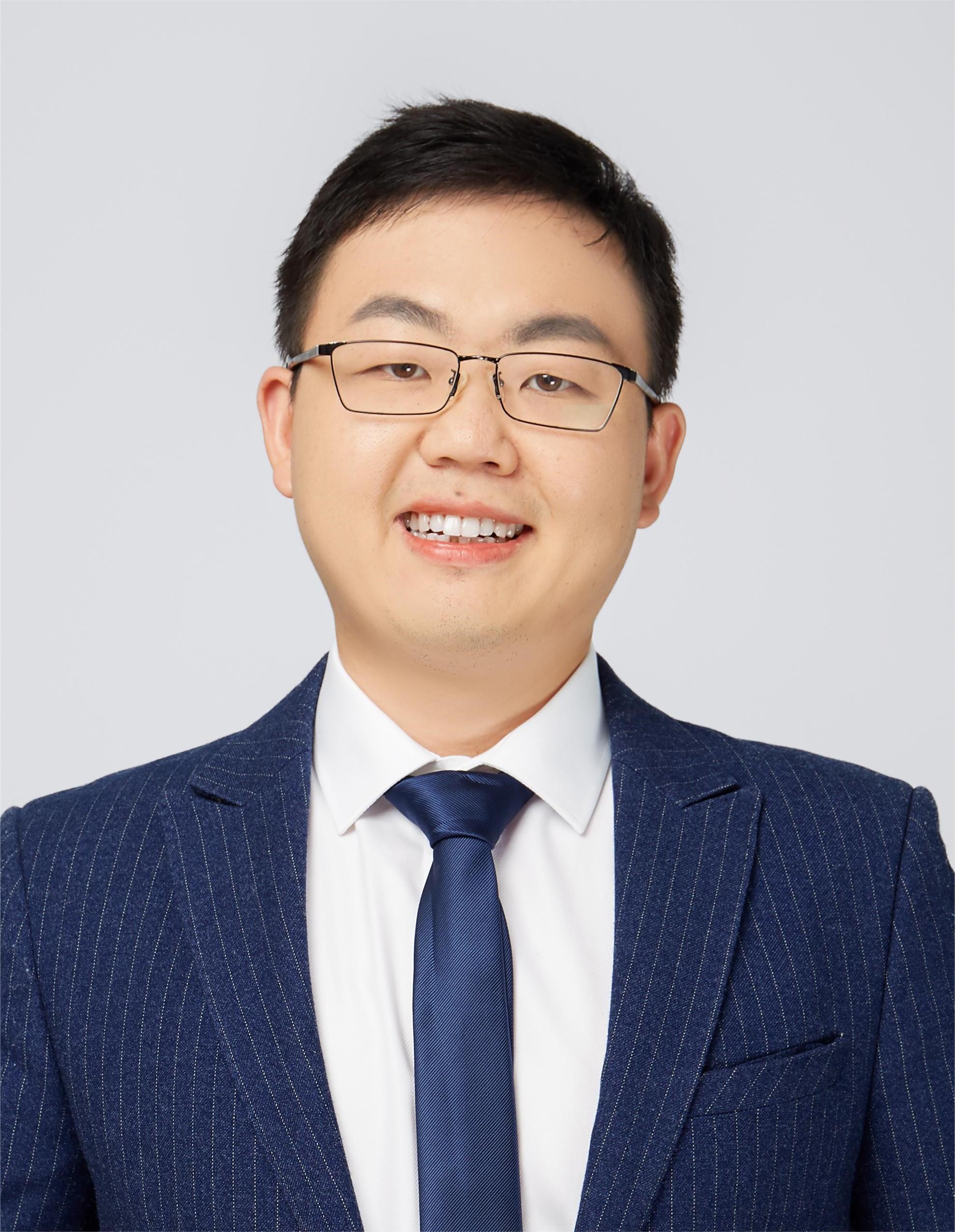}}]{Siyuan Peng} received the BS degree in electronic and communication engineering from Hohai University in 2013, the MS degree in electronic and information engineering from South China University of Technology in 2016 and the PhD degree in Electrical and Electronic Engineering from Nanyang Technological University in 2021 respectively. From 2021 to 2023, he was a Post-Doctoral Researcher in the School of Information Engineering of Guangdong University of Technology. Now he is currently a lecturer in the School of Information Engineering of Guangdong University of Technology, China. His research interests include machine learning, data mining, and signal processing.
\end{IEEEbiography}

\begin{IEEEbiography}
[{\includegraphics[width=1in,height=1.25in,clip,keepaspectratio]{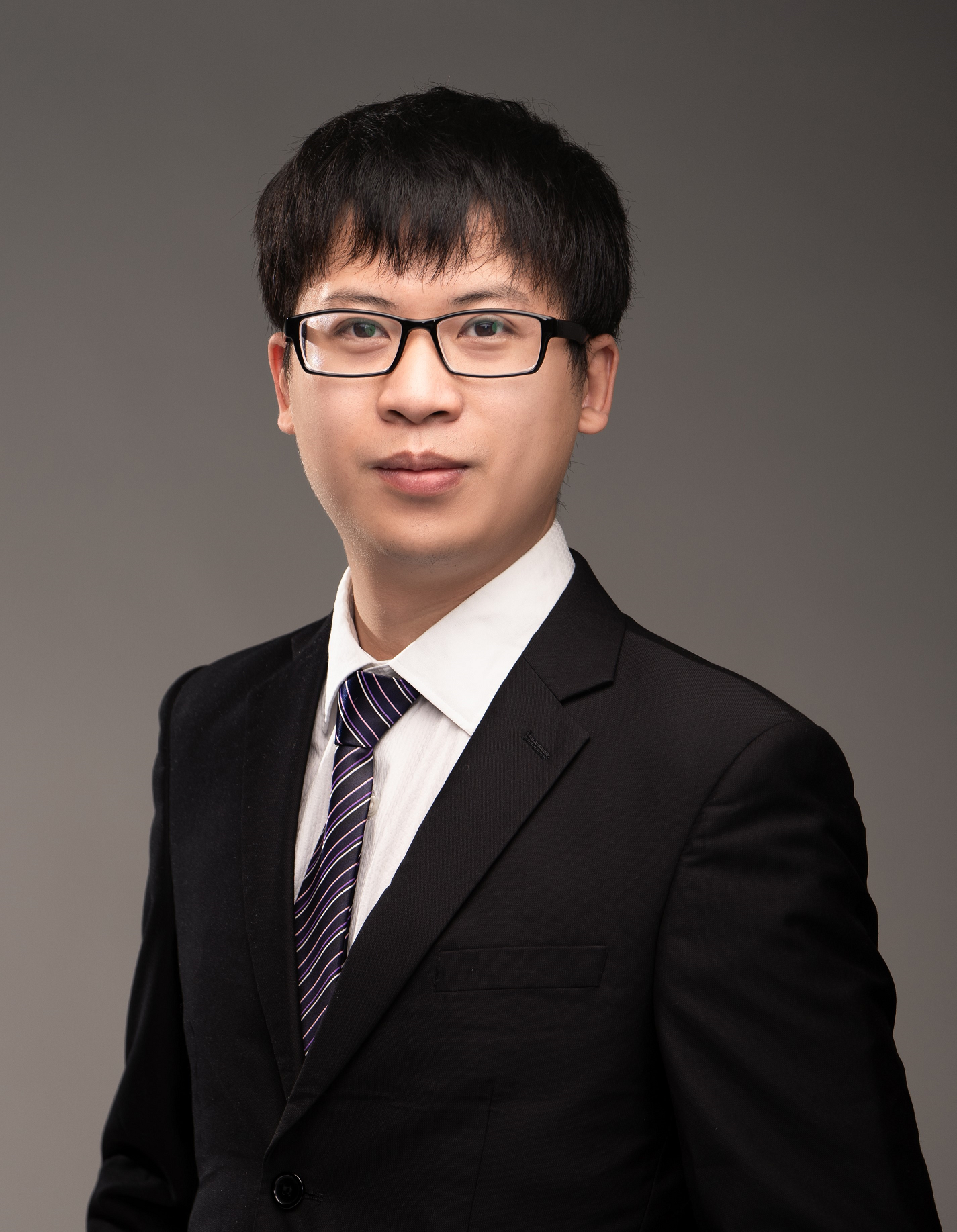}}]{Keze Wang} is nationally recognized as the Distinguished Young Scholars of the National Natural Science Foundation of China (Overseas), currently serving as an Associate Professor at the School of Computer Science, Sun Yat-sen University, and a doctoral supervisor. He holds two Ph.D. degrees, one from Sun Yat-sen University (2017) and another from the Hong Kong Polytechnic University (2019). In 2018, he worked as a postdoctoral researcher at the University of California, Los Angeles, and returned to Sun Yat-sen University in 2021 as part of the ``Hundred Talents Program''. He has focused on reducing deep learning's dependence on training samples and mining valuable information from massive unlabeled data, proposing fundamental learning paradigms, e.g., long-term self-learning and pseudo-label learning mechanisms. This has led to the gradual construction of a theoretical and methodological system for vision computing and reasoning. He has published nearly 30 papers in top-tier journals and conferences, including iScience, T-PAMI, T-NNLS, CVPR, and ICCV, with 12 papers as the first or corresponding author. His works have been cited approximately 2223 times on Google Scholar, and his has three ESI highly cited papers. He holds five patents and has received the 2018 Wu Wenjun AI Science and Technology Award, the 2019 Outstanding Doctoral Dissertation Award, and a nomination for the 2022 AI 2000 Most Influential Scholar Award.
\end{IEEEbiography}

\begin{IEEEbiography}[{\includegraphics[width=1in,clip]{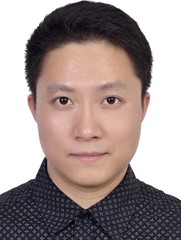}}]{Liang Lin} (Fellow, IEEE) is a full professor at Sun Yat-sen University. From 2008 to 2010, he was a postdoctoral fellow at the University of California, Los Angeles. From 2016--2018, he led the SenseTime R\&D teams to develop cutting-edge and deliverable solutions for computer vision, data analysis and mining, and intelligent robotic systems. He has authored and coauthored more than 100 papers in top-tier academic journals and conferences (e.g., 15 papers in TPAMI and IJCV and 60+ papers in CVPR, ICCV, NIPS, and IJCAI). He has served as an associate editor of IEEE Trans. Human-Machine Systems, The Visual Computer, and Neurocomputing and as an area/session chair for numerous conferences, such as CVPR, ICME, ACCV, and ICMR. He was the recipient of the Annual Best Paper Award by Pattern Recognition (Elsevier) in 2018, the Best Paper Diamond Award at IEEE ICME 2017, the Best Paper Runner-Up Award at ACM NPAR 2010, Google Faculty Award in 2012, the Best Student Paper Award at IEEE ICME 2014, and the Hong Kong Scholars Award in 2014. He is a Fellow of IEEE, IAPR, and IET.
\end{IEEEbiography}

\vfill

\end{document}